\documentclass{article}

\PassOptionsToPackage{numbers, compress}{natbib}

\usepackage[final]{neurips_2020}

\usepackage[utf8]{inputenc} 
\usepackage[T1]{fontenc}    
\usepackage{hyperref}       
\usepackage{url}            
\usepackage{booktabs}       
\usepackage{amsfonts}       
\usepackage{nicefrac}       
\usepackage{microtype}      

\usepackage{amsfonts}
\usepackage{bbm}
\usepackage{mathtools}
\usepackage{amsmath}
\usepackage{amssymb}
\usepackage{multirow}
\usepackage{setspace}
\usepackage{threeparttable}
\usepackage{stfloats}
\usepackage{textcomp}
\usepackage{caption}
\usepackage{chngpage}
\usepackage[inline]{enumitem}
\DeclareMathOperator{\bin}{bin}
\DeclareMathOperator{\sgn}{sign}

\DeclareMathOperator{\bn}{BatchNorm}
\DeclareMathOperator{\var}{Var}
\DeclareMathOperator{\expect}{E}

\DeclareMathOperator*{\argmax}{arg\,max}
\DeclarePairedDelimiter{\nint}\lfloor\rceil
\DeclarePairedDelimiter{\abs}\lvert\rvert
\newcommand{\iquant}[1]{\nint*{\frac{#1}{s}}}
\newcommand{\xquant}{x^{\operatorname{q}}}
\newcommand{\wbin}{W^{\operatorname{bin}}}
\newcommand{\bsat}{b^{\operatorname{SAT}}}
\newcommand{\kbn}{k^{\operatorname{BN}}}
\newcommand{\bbn}{b^{\operatorname{BN}}}
\newcommand{\xbn}{x^{\operatorname{BN}}}
\newcommand{\lcbd}{L^{\operatorname{CBD}}}
\newcommand{\real}{\mathbb{R}}
\newcommand{\intset}{\mathbb{Z}}
\newcommand{\linf}{\ell_{\infty}}
\newcommand{\lone}{\ell_{1}}
\newcommand{\norminf}[1]{\left\lVert#1\right\rVert_{\infty}}
\newcommand{\scinum}[2]{{#1}\mathrm{e}{#2}}

\usepackage[nameinlink,noabbrev]{cleveref}
\renewcommand{\cref}[1]{\Cref{#1}}
\newcommand{\eqnref}[1]{\hyperref[eqn:#1]{(\ref*{eqn:#1})}}

\newenvironment{enuminline}
{\begin{enumerate*}[label=(\roman*),itemjoin={{; }},itemjoin*={{; and }}]}
{\end{enumerate*}}

\usepackage{array}
\newcolumntype{L}[1]{>{\raggedright\let\newline\\\arraybackslash\hspace{0pt}}m{#1}}
\newcolumntype{C}[1]{>{\centering\let\newline\\\arraybackslash\hspace{0pt}}m{#1}}
\newcolumntype{R}[1]{>{\raggedleft\let\newline\\\arraybackslash\hspace{0pt}}m{#1}}

\newenvironment{qanda}{\setlength{\parindent}{0pt}}{\bigskip}
\newcommand{\Q}{\bigskip\bfseries Q: }
\newcommand{\A}{\par\textbf{A:} \normalfont}

\newcommand{\cmpLastBnOursAcc}{55.22\%}

\newcommand{\cmpLastBnFullAcc}{54.74\%}
\newcommand{\cmpLastBnFullLoss}{1.29}
\newcommand{\cmpLastBnNoneAcc}{52.04\%}
\newcommand{\cmpLastBnNoneLoss}{1.35}
\newcommand{\hardtanhCmpBaselinePGD}{19.82\%}
\newcommand{\hardtanhCmpBaselinePGDWithTanh}{23.25\%}
\newcommand{\mnistMlpPrec}{97.14\%}
\newcommand{\mnistMlpNzp}{$18\mathrm{K}$}
\newcommand{\mnistMlpRefPrec}{95.2\%}
\newcommand{\mnistMlpRefNzp}{$20\mathrm{K}$}
\newcommand{\mcsSpeedupMin}{1.35}
\newcommand{\mcsSpeedupAvg}{40.21}
\newcommand{\mcsSpeedupMax}{5104.94}
\newcommand{\mcsSpeedupMidMin}{8.93}
\newcommand{\mcsSpeedupMidMax}{75289.76}
\newcommand{\ternWeightCardBound}{115.3}
\newcommand{\ternWeightCbdCardBound}{4.3}
\newcommand{\ternWeightCbdCardBoundMax}{146.3}
\newcommand{\ternWeightAdvCbdCardBoundMax}{73.1}
\newcommand{\ternWeightStrongCbdCardBound}{3.2}
\newcommand{\ternWeightStrongCbdCardBoundMax}{157.7}
\newcommand{\ternWeightBinMaskCbdCardBound}{3.8}
\newcommand{\ternWeightBinMaskCbdCardBoundMax}{22.3}
\newcommand{\layerSparsityTernary}{9\%\;21\%\;21\%\;28\%\;90\%\;87\%\;41\%}
\newcommand{\layerSparsityBinMask}{84\%\;61\%\;65\%\;61\%\;87\%\;86\%\;70\%}
\newcommand{\layerSparsityTernaryCardBound}{115.3}
\newcommand{\layerSparsityBinMaskCardBound}{119.4}
\newcommand{\solveSpeedupMin}{109.29}
\newcommand{\solveSpeedupMax}{15897.42}
\newcommand{\totalSpeedupMin}{338.28}
\newcommand{\totalSpeedupMax}{2440.65}

\author{ Kai~Jia \\
    MIT CSAIL \\
    32 Vassar St, Cambridge, MA 02139 \\
    \texttt{jiakai@mit.edu} \\
	\And
	Martin Rinard \\
    MIT CSAIL \\
    32 Vassar St, Cambridge, MA 02139 \\
    \texttt{rinard@csail.mit.edu} \\
}

\title{Efficient Exact Verification of Binarized Neural Networks}

\begin{document}

\maketitle

\begin{abstract}
    Concerned with the reliability of neural networks, researchers have
    developed verification techniques to prove their robustness. Most verifiers
    work with real-valued networks. Unfortunately, the exact (complete and
    sound) verifiers face scalability challenges and provide no correctness
    guarantees due to floating point errors. We argue that Binarized Neural
    Networks (BNNs) provide comparable robustness and allow exact and
    significantly more efficient verification. We present a new system, EEV, for
    efficient and exact verification of BNNs. EEV consists of two parts:
    \begin{enuminline}
        \item a novel SAT solver that speeds up BNN verification by natively
            handling the reified cardinality constraints arising in BNN
            encodings
        \item strategies to train solver-friendly robust BNNs by inducing
            balanced layer-wise sparsity and low cardinality bounds, and
            adaptively cancelling the gradients
    \end{enuminline}.
    We demonstrate the effectiveness of EEV by presenting the first exact
    verification results for $\linf$-bounded adversarial robustness of
    nontrivial convolutional BNNs on the MNIST and CIFAR10 datasets. Compared to
    exact verification of real-valued networks of the same architectures on the
    same tasks, EEV verifies BNNs hundreds to thousands of times faster, while
    delivering comparable verifiable accuracy in most cases.
\end{abstract}


\section{Introduction}

Deep Neural Networks (DNNs) have achieved impressive success in many
applications including image understanding, speech recognition, natural language
processing, and game playing~\cite{ian2016deep}. Unfortunately, DNNs exhibit
unexpected or potentially dangerous behavior due to limited robustness~\cite{
szegedy2014intriguing}.

In response, researchers have developed techniques that attempt to verify that a
DNN satisfies a robustness specification~\cite{katz2017reluplex,
ehlers2017formal, tjeng2018evaluating}. However, scalability is a formidable
challenge for exact verifiers~\cite{bunel2018unified}. For example,
\texttt{MIPVerify}~\cite{tjeng2018evaluating} needs hundreds of seconds to
verify the robustness of a network whose inference on CPU takes only a few
milliseconds, even though the network is trained to be solver-friendly~\cite{
xiao2018training}. Additionally, floating point errors make any correctness
guarantees effectively unobtainable for exact verification of nontrivial
real-valued neural networks~\cite{jia2020exploiting}.

Binarized Neural Networks (BNNs) comprise an attractive alternative to
real-valued neural networks. BNNs exhibit significant speed gain and energy
savings during inference~\cite{ hubara2016binarized, rastegari2016xnor,
moss2017high} while achieving competitive accuracy on challenging
datasets~\cite{bethge2019back}. Because of the absence of floating point
arithmetic, BNNs can also support exact verification~\cite{
cheng2018verification, narodytska2018verifying}. To date, however, BNN
verification has exhibited even worse scalability than real-valued neural
network verification~\cite{narodytska2020in}.

We present a new system, EEV, for exact and efficient verification of
BNNs. EEV incorporates a novel SAT solver tailored for BNN verification and new
training strategies that enhance the robustness and verification efficiency of
the trained BNNs. We use EEV to verify the robustness of BNNs against input
perturbations bounded by the $\linf$ norm. Our experimental results show that,
compared to exact verification of robustly trained real-valued networks with the
same architectures, EEV delivers several orders of magnitude faster verification
of BNNs with comparable verifiable accuracy in most cases. This paper makes the
following contributions:

\begin{enumerate}
    \item We incorporate native support for reified cardinality constraints into
        a SAT solver, which improves the performance of BNN verification by more
        than a factor of one hundred compared to an unmodified SAT
        solver~(\cref{sec:minisatcs}).

    \item We identify that sparse weights induced by ternary quantization~\cite{
        narodytska2020in} cause unbalanced sparsity between layers of
        convolutional networks, which leads to high verification complexity
        despite sufficient overall sparsity. We propose a new strategy, BinMask,
        which produces more balanced sparsity. Our system verifies BNNs trained
        with BinMask two to five orders of magnitude faster than BNNs trained
        with ternary quantization~(\cref{sec:binmask}).

    \item We further reduce the verification complexity of BNNs by introducing a
        regularizer that induces lower cardinality bounds, which leads to an
        additional speedup of up to thousands of times~(\cref{sec:cbd}).

    \item We find that directly applying the PGD training algorithm~\cite{
        madry2018towards} does not induce verifiably robust BNNs. We propose
        adaptive gradient cancelling to train robust
        BNNs~(\cref{sec:train-robust}).

    \item We present the first exact verification of robustness against
        $\linf$-norm bounded input perturbations of convolutional BNNs on MNIST
        and CIFAR10, and compare the results with real-valued networks. EEV
        verifies exact robustness of BNNs between \totalSpeedupMin~to
        \totalSpeedupMax~times faster than a state-of-the-art exact verifier on
        real-valued networks with the same architecture, while delivering
        comparable verifiable accuracy in most cases~(\cref{tab:summary}).
\end{enumerate}


\section{Background and related work}

DNN exact (i.e., sound and complete) verification checks whether a DNN satisfies
a specification, which, for example, can require the network to give robust
predictions against input perturbations. Exact verifiers aim to provide either a
proof of satisfaction or a counterexample. Researchers have developed a range of
verification techniques, mostly for real-valued ReLU networks. They typically
formulate the verification as a Satisfiability Modulo Theory (SMT)
problem~\cite{ scheibler2015towards, huang2017safety, katz2017reluplex,
ehlers2017formal} or a Mixed Integer Linear Programming (MILP) problem~\cite{
lomuscio2017approach, cheng2017maximum, fischetti2018deep, dutta2018output,
tjeng2018evaluating}, which can be exponentially slow due to the NP-completeness
of the verification problem\cite{katz2017reluplex}.  Certification (i.e., sound
but incomplete verification) improves scalability at the cost of
completeness~\cite{ wong2017provable, weng2018towards, gehr2018ai2,
zhang2018efficient, raghunathan2018semidefinite, dvijotham2018training,
mirman2018differentiable, singh2019an}. They often explore the idea of
over-approximation to simplify the computation, which causes them unable to
prove or disprove the specification in certain cases. Many of the certification
methods can be unified under a layer-wise convex relaxation framework, but there
seems to be an inherent barrier to tight verification via the relaxation
captured by such a framework~\cite{salman2019convex}.

Until recently, exact verification of DNNs was too computationally expensive to
scale beyond a few hundred neurons. \citet{tjeng2018evaluating} present the
first exact verification of adversarial robustness for convolutional neural
networks (CNNs) on MNIST using an MILP formulation. The verification performance
is further improved by deriving tighter bounds of hidden neurons used in the
MILP formulation~\cite{singh2018robustness}. Training a network with more stable
ReLU neurons also reduces verification complexity and enables evaluating CNN
robustness on the more challenging CIFAR10 dataset~\cite{ xiao2018training}.
However, extant exact verifiers do not attempt to soundly model the floating
point arithmetic in any inference implementation to achieve a manageable
computational complexity~\cite{katz2017reluplex, tjeng2018evaluating}. The
resulting discrepancy between the actual and analyzed inference algorithms
enables constructing adversarial examples for networks that are claimed to be
robust by an exact verifier~\cite{jia2020exploiting}.

Binarized Neural Networks (BNNs)~\cite{hubara2016binarized} constrain their
activations and weights to be binary. Binarization facilitates analysis because
its combinatorial nature enables close interaction with logical reasoning, which
allows a rich set of properties to be encoded in SAT formulas. Examples include
queries on adversarial robustness, Trojan attacks, fairness, network
equivalence, and model counting~\cite{narodytska2018verifying,
baluta2019quantitative}. Analysis techniques for BNNs include efficient
encoding~\cite{shih2019verifying} and exploiting decomposability between neurons
or layers~\cite{ cheng2018verification, khalil2018combinatorial}. However, prior
BNN verifiers have exhibited even worse scalability than real-valued network
verifiers. These scalability challenges have, to date, prevented the evaluation
of any meaningful robustness metric for BNNs~\cite{narodytska2020in}.

Adversarial attack and defense of DNNs is a developing field where most research
focuses on real-valued networks~\cite{ carlini2017towards,athalye2018obfuscated,
madry2018towards, tramer2020adaptive}. BNNs can be attacked by gradient-based
adversaries~\cite{ galloway2018attacking} or specialized solving
algorithms~\cite{ khalil2018combinatorial}. We are the first to design robust
training algorithms specifically for BNNs to induce verifiably robust networks.


\section{Preliminaries}

\label{sec:bnn-def}

The Boolean Satisfiability Problem (SAT) is the problem of deciding whether
there exists a variable assignment to satisfy a given Boolean expression
\cite{biere2009handbook}.  We consider Boolean expressions in Conjunctive Normal
Form (CNF) defined over a set of Boolean variables ${x_1, \cdots, x_n}$. A CNF
$e$ is a conjunction of a set of \emph{clauses}: $e = c_1 \land \cdots \land
c_m$, where each clause $c_i$ is a disjunction of some \emph{literals} $c_i =
l_{i1} \lor \cdots \lor l_{is_i}$, and a literal $l_{ij}$ is either a variable
or its negation: $l_{ij} = x_k$ or $l_{ij} = \lnot x_k$. Despite the well known
fact that 3-SAT is NP-complete \cite{cook1971complexity}, efficient heuristics
have been developed to scale SAT solvers to handle industrial level
problems~\cite{balyo2017sat}.

Binarized Neural Networks (BNNs)~\cite{hubara2016binarized} quantize their
weights and activations to be binary. The BNN is proposed as a method to reduce
the computation burden and speed up inference and possibly also
training~\cite{rastegari2016xnor, zhou2016dorefa, jacob2018quantization}.

The basic building block of a BNN is a \emph{linear-BatchNorm-binarize}
module that maps an input tensor $x\in\{0,1\}^n$ to an output tensor
$y\in\{0,1\}^m$ with a weight parameter $W \in \real^{m\times n}$ and also
trainable parameters $\gamma\in\real^m$ and $\beta\in\real^m$ in the batch
normalization~\cite{ioffe2015batch}:
\begin{align}
    y = \bin_{act}(\bn(\bin_{w}(W) x)) \label{eqn:bnn-layer-train}
\end{align}
where:
\begin{align*}
    \bin_{w}(W) &= \sgn(W) \in\{-1,1\}^{m\times n} \\
    \sgn(x) &= \left\{ \begin{array}{ll}
        1 & \text{ if } x \ge 0 \\
        -1 & \text{ otherwise}
    \end{array}\right. \\
    \bn(x) &= \gamma \odot \frac{x-\expect[x]}{\sqrt{\var[x]+\epsilon}} +
        \beta \\
    & \text{where } \epsilon = \scinum{1}{-5} \text{, and $\odot$ denotes
        element-wise multiplication} \\
    \bin_{act}(x) &= (x \ge 0) = (\sgn(x)+1)/2 \in\{0, 1\}^m
\end{align*}

Note that we use $\{0,1\}$ for activations rather than $\{-1,1\}$ that is
commonly adopted in the literature~\cite{hubara2016binarized,
narodytska2018verifying}. Although both representations have identical
representation capability because they are linear transformations of each other,
our representation simplifies both the SAT encoding process and the zero padding
of convolutional layers. The weight $\bin_w(W)$ can take zero values in a sparse
network.

\noindent{\bf First layer:} The first layer of a BNN is usually applied on floating
point inputs or fixed-point numbers~\cite{hubara2016binarized}. However,
encoding floating-point or integer arithmetic in SAT typically incurs high
complexity. To simplify the verification process, we quantize the inputs:
\begin{align}
    \xquant = \iquant{x} \cdot s \label{eqn:input-quant}
\end{align}
where $x\in\real_{[0,1]}^n$ is the real-valued input, $\xquant$ is the quantized
input to be fed into the BNN, and $s$ is the quantization step size which can be
set to $s=1/255$ for emulating 8-bit fixed point values, or $2\epsilon$ for
adversarial training with a $\linf$ perturbation bound of $\epsilon$. Since a
robust network should be invariant to perturbations within $[x-\epsilon,
x+\epsilon]$, we expect the quantization with $s=2\epsilon$ not to discard
information useful for robust classification, which is confirmed by checking
that a few choices of the quantization step do not noticeably affect test
accuracy.

\noindent{\bf Last layer:} We consider the layer before softmax as the last layer
of the BNN, which outputs a logits vector that can be interpreted as the
classification score. We remove the $\bin_{act}$ in \eqnref{bnn-layer-train} to
obtain a real-valued score. To facilitate the SAT conversion, we also restrict
the variance statistics and the scale parameter $\gamma$ in $\bn(\cdot)$ of the
last layer to be scalars computed on the whole feature map rather than
per-channel statistics. In practice, such a restriction on the final batch
normalization does not affect network performance. For the undefended
\texttt{conv-small} network trained on CIFAR10, using the original per-channel
batch normalization in the last layer achieves $\cmpLastBnFullAcc$ accuracy and
a negative log-likelihood loss of $\cmpLastBnFullLoss$ on the test set, while
the restricted network delivers $\cmpLastBnOursAcc$ test accuracy and the same
loss.  However, if the batch normalization is removed from the last layer, the
test accuracy drops to $\cmpLastBnNoneAcc$ and the loss increases to
$\cmpLastBnNoneLoss$.


\section{Combinatorial analysis of BNNs}
\subsection{Encoding BNNs with reified cardinality constraints}

We outline the encoding of a BNN and its robustness specification with respect
to $\linf$-norm bounded perturbations around a given input as a set of Boolean
clauses. The details are presented in the supplementary material.

Let $\wbin = \bin_w(W)$ denote the sparse weights. The inference computation of
a linear-BatchNorm-binarize module described in \cref{sec:bnn-def} can be
formulated as:
\begin{align}
    y = \bin_{act}(\kbn \odot (\wbin x) + \bbn)
    \label{eqn:bnn-layer-inf-lin}
\end{align}
where $\kbn$ and $\bbn$ are constants derived from batch normalization weights
and statistics. Since values of $\wbin$ fall in $\{-1,0,1\}$, we can rewrite
\eqnref{bnn-layer-inf-lin} as a set of \emph{reified cardinality constraints}
between $x$ and $y_i$:
\begin{align}
    y_i = \left(\sum_{j=1}^n l_{ij}(x_j) \gtreqless
        \left[b_i(\kbn,\wbin,\bbn)\right] \right)
    \label{eqn:bnn-rcc-form}
\end{align}
where $l_{ij}(x)$ takes one value of $\{0, x, \lnot x\}$ depending on
$\wbin_{ij}$, $b(\kbn,\wbin,\bbn)$ is a constant derived from the weights,
$[\cdot]$ acts as flooring or ceiling according to the sign of $\kbn$, and
$\gtreqless$ acts as $\geq$ or $\leq$ according to the sign of the $\kbn$. In
the encoding we treat \texttt{TRUE} and $1$ equivalently, \texttt{FALSE} and $0$
equivalently, and the comparator $\gtreqless$ produces a value in $\{0,1\}$.

Let $x$ denote the input image, $y$ denote the logits vector of the last layer,
and $c$ denote the target class number. To evaluate robustness under
$\linf$-norm bounded perturbations, we need to encode the input perturbation
space $\norminf{x-x_0} \le \epsilon$ and the untargeted attack goal $\lor_{i\neq
c}\left(y_i - y_c > 0\right)$. Recall that inputs are quantized by $\xquant =
\iquant{x}\cdot s$, which enables us to only encode the integer interval of
allowed $\iquant{x}$ values by merging the multiplier $s$ into $\kbn$ of the
first layer. Encoding the constraint $v=\iquant{x}\in\intset\cap[a,b]$ is
achieved by introducing $b-a$ auxiliary Boolean variables
$\{t_1,\cdots,t_{b-a}\}$ and assigning $v = a+\sum_{i=1}^{b-a}t_i$. We further
reduce the search space by enforcing the thermometer
encoding~\cite{buckman2018thermometer} on $\{t_1,\cdots,t_{b-a}\}$ so that the
encoding of each value of $v$ is unique, via adding clauses $t_i \lor \lnot t_j$
for $1 \le i < j \le b-a$.  The clause $y_i - y_c > 0$ in the untargeted attack
goal is converted into a reified cardinality constraint similarly to the layer
encoding of \eqnref{bnn-layer-inf-lin} and \eqnref{bnn-rcc-form}.

Although the BNN is a special case of real-valued NNs, we still need to develop
BNN-specific encodings because existing verifiers for real-valued NNs are unable
to handle the $\sgn(\cdot)$ nonlinearity. It has also been shown that a
straightforward MILP or ILP encoding for a BNN results in slower solving
compared to the SAT encoding~\cite{ narodytska2018verifying}. Note that
computing the SAT encodings of BNNs only requires a linear scan of the network
weights and architecture, and a verifier can cache the encoded logical
constraints of the same network to be used for different inputs. Computing BNN
encodings is easier and faster compared to computing the MILP encodings for
real-valued networks that usually need to estimate the bounds of hidden neurons
for each network input.

Compared to the SAT encoding in previous BNN verification research~\cite{
narodytska2018verifying, narodytska2020in}, our network design and the
corresponding encoding enjoy three benefits:
\begin{enuminline}
    \item By replacing $\{-1,1\}$ activations with
        $\{0,1\}$ activations that directly correspond to Boolean values, the
        encoding is simplified and zero padding in convolutional layers is
        trivially supported, while \citet{narodytska2020in} only evaluate fully
        connected networks
    \item More information in the input is retained by quantization rather than
        binarization
    \item Test accuracy is improved by incorporating batch normalization in the
        last layer
\end{enuminline}.

\subsection{MiniSatCS: an efficient SAT solver for reified cardinality
constraints}
\label{sec:minisatcs}

Modern SAT solvers typically build on the Conflict-Driven Clause Learning (CDCL)
algorithm~\cite{marques2009conflict} to search for a solution of a set of
disjunctive clauses. The strategy is to reduce the search space by learning new
clauses from conflicts. There are three key procedures in this algorithm:
\begin{enumerate}
    \item \emph{Branching:} Pick an undecided variable and assign a value to it.
        The order of branching is usually decided by heuristics like
        VSIDS~\cite{moskewicz2001chaff}.
    \item \emph{Propagation:} Unit clauses are detected to infer undecided
        variables given the current variable assignment. A \emph{unit clause}
        contains only one unassigned literal: if there is a clause $c = l_1 \lor
        \dots \lor l_n$ in the database and $l_1, \dots, l_{n-1}$ all evaluate
        to \texttt{FALSE} under current assignment, then $l_n$ must be
        \texttt{TRUE} so the whole clause could be satisfied.
    \item \emph{Clause learning:} When a clause evaluates to \texttt{FALSE}
        after a propagation step, the algorithm tracks the history of
        propagations that leads to the conflict, and determines the set of
        branching choices $x_1 = v_1, \cdots, x_k = v_k$ that are ultimately
        responsible for this conflict, where $x_i$ is a branching variable and
        $v_i \in\{ \texttt{TRUE},\,\texttt{FALSE}\}$ is the branching decision.
        A new clause $(x_1 = \lnot v_1) \lor \cdots \lor (x_k = \lnot v_k)$ is
        inserted into the database of learned clauses to facilitate future
        propagation.
\end{enumerate}

We extend a CDCL-based SAT solver to natively handle reified cardinality
constraints. We compare our technique with two existing approaches for solving a
Boolean system with such constraints:
\begin{itemize}
    \item Deploy an encoding algorithm to convert each reified cardinality
        constraint into a set of disjunctive clauses to be solved by an
        off-the-shelf SAT solver. The sequential counters encoder~\cite{
        sinz2005towards}, which needs $O(nb)$ auxiliary variables and clauses
        for a constraint with $n$ literals and a bound of $b$, has been used by
        previous BNN verifiers~\cite{narodytska2018verifying,narodytska2020in}.
        An arguably more efficient encoder, the cardinality
        networks~\cite{asin2009cardinality} that need $O(n \log^2 b)$ auxiliary
        variables and clauses, has been adopted for learning transition models
        from BNNs for planning~\cite{say2020compact}.
    \item Solve the pseudo-Boolean constraints derived from the original
        constraints. Pseudo-Boolean constraints allow Boolean literals to be
        multiplied by integer weights. A reified cardinality constraint $y =
        (\sum_{i=1}^n l_i \ge b)$ can be encoded by two linear pseudo-Boolean
        constraints:
        \begin{align}
            \left\{
                \begin{array}{lr}
                    \sum_{i=1}^n l_i + b\cdot \lnot y \ge b &
                        \text{  which encodes }
                        \sum_{i=1}^nl_i < b \implies y=0 \\
                    (n-b+1)\cdot y - 1 \ge \sum_{i=1}^n l_i - b &
                        \text{  which encodes }
                        \sum_{i=1}^nl_i \ge b \implies y=1 \\
                \end{array}
            \right.
            \label{eqn:pb-enc}
        \end{align}
        We are unaware of previous use of this encoding in BNN analysis. We
        include this comparison as one potential alternative to our direct
        support of reified cardinality constraints.
\end{itemize}

Our extension builds on the observation that the CDCL framework can be
generalized to handle clauses not in the disjunctive form, as long as each
clause permits inferring values of undecided variables. This idea has been
explored in the literature to extend SAT solvers to domain-specific
problems~\cite{mate2009extending, liffiton2012cardinality, ganesh2012lynx}. A
key component of our extension is to efficiently handle the propagation through
a reified cardinality constraint $y=\left(\sum_{i=1}^nl_i \le b\right)$, which
contains two cases:
\begin{itemize}
    \item \emph{Operand-inferring:} If $y$ is known and enough of the $\{l_i\}$
        are known, then the remaining $\{l_i\}$ can be inferred. For example, if
        $y$ is known to be true and there are already $b$ literals in $\{l_i\}$
        known to be true, then the other literals must be false.
    \item \emph{Target-inferring:} If enough of the $\{l_i\}$ are known, then
        $y$ can be inferred. For example, if the number of false literals in
        $\{l_i\}$ reaches $n-b$, then $y$ can be inferred to be true.
\end{itemize}

We present \texttt{MiniSatCS}, a novel system that efficiently and natively
handles reified cardinality constraints based on the above ideas.
\texttt{MiniSatCS} extends \texttt{MiniSat 2.2} which is a highly optimized and
minimalistic SAT solver based on years of SAT solving research~\cite{
een2003extensible}. We extend the clause data structure to represent both the
reified less-equal cardinality constraints and disjunction constraints.
\texttt{MiniSatCS} maintains counters that keep the current number of known true
or false literals for each clause. The counters are updated when related
variable assignment changes by a notification system similar to watched
literals~\cite{ moskewicz2001chaff}, so repeated scanning of clauses is avoided
during search for unit clauses.  We use random polarity and turned off the phase
saving heuristics~\cite{ pipatsrisawat2007lightweight} in the solver since it is
faster for BNN verification.

\begin{figure}[h]
    \centering
    \includegraphics[width=0.95\columnwidth]{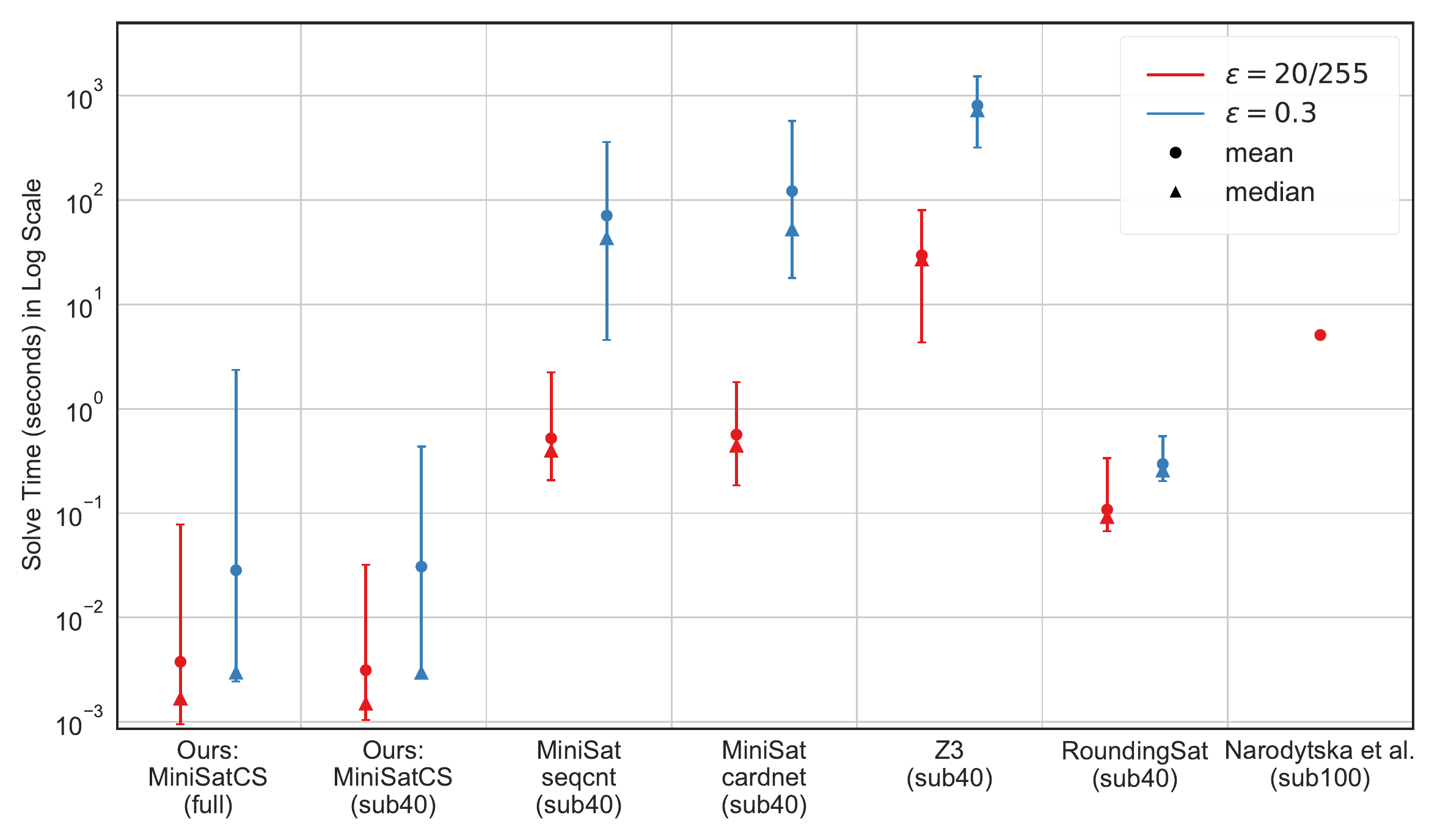}
    \scriptsize
    \caption{Performance comparison on searching adversarial inputs for an
        undefended \texttt{MNIST-MLP} network. The graph plots the
        min/median/mean/max times of each solver on the indicated set of
        benchmarks, with a one hour time limit. The \texttt{(sub40)} or
        \texttt{(sub100)} suffix indicates testing on a subset containing 40 or
        100 MNIST test images, and \texttt{(full)} indicates testing on the
        complete test set.  Note that the solving time is in log scale, and our
        method exhibits especially low median solving times. Our
        \texttt{MNIST-MLP} network achieves \mnistMlpPrec~test accuracy with
        \mnistMlpNzp~non-zero parameters and an input quantization step $s=0.1$,
        compared to \mnistMlpRefPrec~accuracy with \mnistMlpRefNzp~non-zero
        parameters and binarized inputs reported by \citet{narodytska2020in}.
    }
    \label{fig:cmp-minisat}
\end{figure}

We compare the performance of \texttt{MiniSatCS} on the \texttt{MNIST-MLP}
network (defined in \cref{sec:expr}) against \texttt{MiniSat 2.2} with two
encoding choices, two other state-of-the-art solvers, and prior BNN
verification:
\begin{enuminline}
    \item \texttt{MiniSat-seqcnt} that applies \texttt{MiniSat 2.2} on the
        sequential counters encoding~\cite{sinz2005towards}
    \item \texttt{MiniSat-cardnet} that applies \texttt{MiniSat 2.2} on the
        cardinality networks encoding~\cite{asin2009cardinality}
    \item a general SMT solver \texttt{Z3}~\cite{de2008z3} that supports
        pseudo-Boolean logic via on-demand compilation into sorting circuits
    \item \texttt{RoundingSat}~\cite{devriendt2020learn} with the encoding
        outlined in \eqnref{pb-enc}, which is a state-of-the-art pseudo-Boolean
        solver accelerated by incremental linear program solving, with native
        handling of linear pseudo-Boolean constraints
    \item the end-to-end solving time for verifying the same BNN architecture
        reported in prior state-of-the-art BNN verification resarch~\cite{
        narodytska2020in}, which uses the Glucose SAT solver~\cite{
        audemard2018glucose} with the sequential counters encoding
\end{enuminline}.
We present the solving time details in \cref{fig:cmp-minisat}. Our results show
that \texttt{MiniSatCS} is several orders of magnitude faster than other solvers
and the prior state-of-the-art result of BNN verification on the same
architecture. We present more results in the supplementary material.


\section{Training solver-friendly BNNs}
\subsection{BinMask: balanced weight sparsifying}
\label{sec:binmask}

Sparse weights are known to facilitate DNN verification~\cite{
tjeng2018evaluating,xiao2018training,narodytska2020in}. BNNs are commonly
sparsified by ternary quantization~\cite{narodytska2020in} that sets $\bin_w(W)
= 0$ when $\abs{W} < T$. However, ternary quantization suffers from two
drawbacks:
\begin{enuminline}
    \item The threshold $T$ and the penalty coefficient for $\lone$
        regularization are two coupled parameters that need tuning
    \item For convolutional networks, the sparsity of convolutional layers is
        usually lower than that of fully connected layers, which has also been
        observed during pruning real-valued networks~\cite{han2015learning}.
\end{enuminline}
The low sparsity of the convolutional layers limits verification speedup because
those layers reuse the weights and lower weight sparsity leads to higher
verification complexity. While it is possible to prune each layer with a fixed
rate and retrain the network iteratively~\cite{frankle2018the}, such methods are
especially costly for adversarial training.

We propose to induce more balanced layer-wise sparsity by decoupling the
optimization of weight value and weight sparsity. Our method also eliminates the
threshold hyperparameter. A binary mask is applied on the weights, so that in
the training process, the weights become gradually more sparse by masking out
less important weights and the sign of unmasked weights can still be optimized.
More formally, for each weight $W$ we introduce a new mask weight $M_W$ to be
optimized independently of $W$ and replace the definition of $\bin_w$ in
\eqnref{bnn-layer-train} by:
\begin{align}
    \bin_w(W) = \sgn(W) \odot \frac{\sgn(M_W)+1}{2} \label{eqn:binmask}
\end{align}

The mask $M_W$ is initialized from a folded normal distribution (i.e., absolute
value of Gaussian) so that the training starts from a dense network. Weight
decay is applied on the positive elements in the mask $M_W$. We call our
formulation \emph{BinMask} and present results of an empirical comparison with
ternary quantization in \cref{tab:cmp-binmask}. Although the two methods achieve
similar total sparsity and test accuracy, BinMask induces more balanced
layer-wise sparsity, and consequently its verification is thousands of times
faster. We present additional experiments and further discussions in the
supplementary material.

\begin{table}[t]
    \centering
    \scriptsize
    \caption{Comparing BinMask and ternary quantization on undefended
        \texttt{conv-small} networks. While both methods produce similar total
        sparsity, the more balanced layer-wise sparsity of BinMask results in
        faster verification. Total sparsity is the proportion of zero parameters
        in the whole network, which is largely determined by the sparsity of the
        third layer --- a fully connected layer with a large weight matrix.
        Solve time is measured by applying \texttt{MiniSatCS} on $40$ randomly
        chosen test images with a one hour time limit.
    }
    \vspace{1em}
    \label{tab:cmp-binmask}
    \begin{tabular}{lrrrr}
\toprule
{} & \multicolumn{2}{c}{MNIST \hspace{1em} $\epsilon=0.1$} & \multicolumn{2}{c}{CIFAR10 \hspace{1em} $\epsilon=2/255$} \\ \cmidrule(lr){2-3} \cmidrule(lr){4-5}
Sparsifier&                           Ternary &              BinMask &                               Ternary &              BinMask \\
\midrule
Total Sparsity          &                              81\% &                 84\% &                                  82\% &                 79\% \\
Layer-wise Sparsity     &               16\% 40\% 84\% 36\% &  91\% 90\% 83\% 92\% &                   16\% 48\% 84\% 23\% &  84\% 85\% 79\% 87\% \\
Mean Solve Time (sub40) &                           756.288 &                0.002 &                                 0.267 &                0.003 \\
Max Solve Time (sub40)  &                          3600.001 &                0.007 &                                 5.707 &                0.005 \\
Natural Test Accuracy   &                           97.59\% &              97.35\% &                               54.78\% &              55.22\% \\
\bottomrule
\end{tabular}

\end{table}

\subsection{Cardinality bound decay}
\label{sec:cbd}

\begin{table}[t]
    \scriptsize
    \centering
    \caption{Effect of Cardinality Bound Decay (CBD) on adversarially trained
        \texttt{conv-large} networks. The CBD loss effectively reduces
        cardinality bounds, resulting in significant verification speedup. The
        results suggest that it also improves robustness, perhaps by
        regularizing model capacity. Solve time is measured by applying
        \texttt{MiniSatCS} on $40$ randomly chosen test images with a one hour
        time limit. Verifiable accuracy is evaluated on the complete test set
        without time limit for networks that can be verified within one second
        per case on average.
    }
    \vspace{1em}
    \label{tab:cmp-cbd}
    \bgroup
        \setlength{\tabcolsep}{1.1ex}
        \begin{tabular}{lrrrrrrrr}
\toprule
{} & \multicolumn{4}{c}{MNIST \hspace{1em} $\epsilon=0.3$} & \multicolumn{4}{c}{CIFAR10 \hspace{1em} $\epsilon=8/255$} \\ \cmidrule(lr){2-5} \cmidrule(lr){6-9}
CBD Loss Penalty ($\eta$)&                               $0$ & $\scinum{ 1 }{ -5 }$ & $\scinum{ 1 }{ -4 }$ & $\scinum{ 5 }{ -4 }$ &                                   $0$ & $\scinum{ 1 }{ -5 }$ & $\scinum{ 1 }{ -4 }$ & $\scinum{ 5 }{ -4 }$ \\
\midrule
Mean / Max Card Bound        &                     148.3~/~364.0 &           4.3~/~15.1 &            3.2~/~9.0 &            2.7~/~6.8 &                         123.8~/~312.0 &            3.1~/~7.4 &            2.7~/~5.9 &            2.1~/~6.4 \\
Mean Solve Time (sub40)      &                          2442.698 &               32.776 &                0.009 &                0.005 &                               206.037 &                0.010 &                0.009 &                0.009 \\
Max Solve Time (sub40)       &                          3600.006 &             1287.739 &                0.040 &                0.012 &                              3600.001 &                0.019 &                0.013 &                0.014 \\
Verifiable Accuracy          &                                 - &                    - &              69.04\% &              72.48\% &                                     - &              19.28\% &              18.81\% &              20.08\% \\
Natural Test Accuracy        &                           98.88\% &              97.37\% &              96.97\% &              96.26\% &                               53.91\% &              40.80\% &              38.75\% &              35.17\% \\
PGD Accuracy                 &                           89.23\% &              87.60\% &              87.82\% &              87.82\% &                               15.32\% &              27.06\% &              26.22\% &              24.69\% \\
First Layer / Total Sparsity &                       85\%~/~82\% &          86\%~/~89\% &          83\%~/~89\% &          82\%~/~87\% &                           95\%~/~88\% &          95\%~/~94\% &          94\%~/~87\% &          89\%~/~90\% \\
\bottomrule
\end{tabular}

    \egroup
\end{table}

While BinMask alone sparsifies the small network enough to be efficiently
verified, it is not sufficient for a larger network. To further reduce
verification complexity, we revisit the reified cardinality constraint $y =
\left(\sum_{i=1}^n l_i \leq b\right)$ and note the following facts:
\begin{enumerate}
    \item The sequential counters encoder~\cite{sinz2005towards} that converts
        the constraint into CNF needs $O(nb)$ auxiliary variables and clauses.
        Thus smaller $b$ produces a simpler encoding.

    \item \texttt{MiniSatCS} can infer $y$ to be false once the number of true
        literals in $\{l_i\}$ exceeds $b$, and a smaller $b$ increases the
        likelihood of this inference.

    \item If the literals $\{l_i\}$ are drawn from independent symmetrical
        Bernoulli distributions, then the entropy of $y$ is a symmetrical
        concave function with respect to $b$ which is maximized when
        $b=\frac{n}{2}$.  Therefore the further $b$ deviates from $\frac{n}{2}$,
        the more predictable $y$ becomes.
\end{enumerate}

We are thus motivated to regularize the bound in reified cardinality
constraints to reduce verification complexity. We propose a Cardinality Bound
Decay (CBD) loss to achieve this goal, by adding an $\lone$ penalty on the bias
terms in~\eqnref{bnn-rcc-form} that exceed a threshold $\tau$:
\begin{align}
    \lcbd &= \eta \max\left(b(\kbn,\wbin,\bbn) - \tau,\, 0 \right)
    \label{eqn:loss-cbd}
\end{align}
where $\eta$ is a coefficient controlling the strength of this loss. We set
$\tau=5$ in our experiments because it is an empirically good choice among a few
values that enables fine control of the accuracy-speed tradeoff by tuning
$\eta$. Note that since $\sum_{i=1}^n l_i \le b$ is equivalent to $\sum_{i=1}^n
\lnot l_i \ge n - b$, we only need to minimize $b$ rather than maximize
$\abs*{b-\frac{n}{2}}$ in this loss. The removal of $n$ results in a simpler
formulation. We present an empirical evaluation of the CBD loss in
\cref{tab:cmp-cbd}, which shows that the CBD loss reduces the cardinality bounds
and verification complexity significantly. Interestingly, the results also
suggest that the CBD loss serves as an effective regularizer which favors model
robustness over natural test accuracy. Note that the cardinality bound is
another measurement of verification complexity different from sparsity, since
they are not strongly correlated.

\section{Training robust BNNs with adaptive gradient cancelling}
\label{sec:train-robust}

Although the $\sgn$ function has zero gradient almost everywhere, a BNN can
still be trained using gradient based optimizers by adopting the
straight-through estimator~\cite{bengio2013estimating} which treats the $\sgn$
function as an identity function during backpropagation. This approach also
enables training robust BNNs with a projected gradient descent~(PGD) adversary,
similar to training robust real-valued networks~\cite{madry2018towards}.

For a sign function quantization $q = \sgn(r)$, empirical results suggest that
gradient cancelling via backpropagating with $g_r = g_q \mathbbm{1}_{|r| \le 1}$
improves both training and test accuracy~\cite{hubara2016binarized}, where $g_q$
and $g_r$ are the gradients of the loss with respect to $q$ and $r$
respectively. Gradient cancelling can be seen as computing the gradient as if
the forward propagation is hard tanh $q = \operatorname{htanh}(r) =
\operatorname{clip}(r, -1, 1)$.

Using the hard tanh gradient cancelling in the PGD adversarial training produces
networks with a high PGD test accuracy, but the true adversarial accuracy
evaluated by an exact verifier still remains low. This suggests that the PGD
attack becomes ineffective. To improve the robustness, we instead propose to use
standard tanh (i.e., backpropagating with $g_r = (1- \tanh^2 r)g_q$) to provide
more effective gradients for both training and the PGD attack. Using tanh
gradient cancelling in PGD improves the attack success rate on a
\texttt{conv-small} network from \hardtanhCmpBaselinePGD~to
\hardtanhCmpBaselinePGDWithTanh.

We generalize the idea and propose adaptive gradient cancelling: $g_r = (1-
\tanh^2 (\alpha r))g_q$, where $\alpha$ is a global parameter to control the
strength of gradient cancelling. Motivated by the intuition that a stronger PGD
attack indicates better gradients, $\alpha$ is chosen to maximize PGD attack
success rate at the beginning of each epoch. The robustness of the
\texttt{conv-large} networks is further improved by generating adversarial
examples with the \texttt{MiniSatCS} verifier for images where the PGD attack
fails in the last ten epochs. \cref{tab:cmp-tanh} compares different gradient
computing methods.

\begin{table}[h]
    \scriptsize
    \centering
    \caption{Comparing gradient computing methods for adversarial training on
        CIFAR10 with the \texttt{conv-large} network and $\epsilon=8/255$. The
        PGD accuracy is evaluated on the test set with the same gradient
        computing as in training. The verifiable accuracy is the exact
        adversarial robustness. Tanh gradient cancelling improves all the
        metrics, and adaptive gradient cancelling reduces the gap between PGD
        accuracy and verifiable accuracy.
    }
    \label{tab:cmp-tanh}
    \vspace{1em}
    \begin{tabular}{lrrrr}
\toprule
{} & hard tanh &     tanh & adaptive & adaptive + verifier adv \\
\midrule
Natural Test Accuracy &   35.42\% &  38.79\% &  35.17\% &                 35.00\% \\
PGD Accuracy          &   22.79\% &  24.98\% &  24.69\% &                 26.41\% \\
Verifiable Accuracy   &   11.13\% &  14.70\% &  20.08\% &                 22.55\% \\
\bottomrule
\end{tabular}

\end{table}


\section{Experiments}
\label{sec:expr}

We adopt three network architectures from the literature for the evaluation of
EEV. \texttt{MNIST-MLP} is a binarized multilayer perceptron with hidden
layers having $[500, 300, 200, 100, 10]$ units~\cite{ narodytska2020in}.
\texttt{Conv-small} is a network with two convolutional layers of $16$
and $32$ channels, followed by two fully connected layers with $100$ and
$10$ units. The convolutional layers have $4\times 4$ filters and
$2\times 2$ stride with a padding of $1$. \texttt{Conv-large} has four
convolutional layers with $[32, 32, 64, 64]$ channels and $[3\times 3,\, 4\times
4,\, 3\times3,\, 4\times 4]$ spatial filters respectively, where each layer has
a padding of $1$ and the larger convolutions have a stride of $2\times 2$. They
are followed by three fully connected layers with $[512, 512, 10]$ output units.
The convolutional BNNs have the same architectures as in \cite{xiao2018training}
except that we binarize the network. We present details of the experimental
settings in the supplementary material. Our source code is available at
\url{https://github.com/jia-kai/eevbnn}.

\paragraph{Verifying adversarial robustness:} We evaluate the performance of EEV
on the task of verifying robustness of BNNs against $\linf$-bounded input
perturbations on the MNIST and CIFAR10 benchmarks. The networks are trained with
adaptive gradient cancelling. We present our results in \cref{tab:summary}. It
shows that EEV achieves solving times \solveSpeedupMin~to \solveSpeedupMax~times
faster (and total times \totalSpeedupMin~to \totalSpeedupMax~times faster) than
state-of-the-art exact verification of real-valued networks~\cite{
xiao2018training} with comparable verifiable accuracy in most cases.

\begin{table}[t]
    \scriptsize
    \centering
    \caption{Results of verifying adversarial robustness. EEV verifies BNNs
        significantly faster with comparable verfiable accuracy in most cases.}
    \label{tab:summary}
    \vspace{1em}
    \begin{threeparttable}
        \begin{tabular}{llrrrrrrr}
\toprule
                                 &                                     & \multicolumn{3}{c}{Mean Time (s)} & \multicolumn{3}{c}{Accuracy} & Timeout \\ \cmidrule(lr){3-5} \cmidrule(lr){6-8}
&                                     &         Build &   Solve &   Total & Verifiable &  Natural &      PGD &       ~ \\
\midrule
\multirow{4}{4.5em}{MNIST $\epsilon=0.1$} & EEV S &        0.0158 &  0.0004 &  0.0162 &    89.29\% &  97.44\% &  93.47\% &       0 \\
                                 & EEV L &        0.1090 &  0.0025 &  0.1115 &    91.68\% &  97.46\% &  95.47\% &       0 \\
                                 & \citet{xiao2018training} S &          4.98 &    0.49 &    5.47 &    94.33\% &  98.68\% &  95.13\% &  0.05\% \\
                                 & \citet{xiao2018training} L\tnote{*} &        156.74 &    0.27 &  157.01 &     95.6\% &  98.95\% &  96.58\% &       0 \\
\cmidrule(lr){1-9}
\multirow{4}{4.5em}{MNIST $\epsilon=0.3$} & EEV S &        0.0140 &  0.0006 &  0.0146 &    66.42\% &  94.31\% &  80.70\% &       0 \\
                                 & EEV L &        0.1140 &  0.0039 &  0.1179 &    77.59\% &  96.36\% &  87.90\% &       0 \\
                                 & \citet{xiao2018training} S &          4.34 &    2.78 &    7.12 &    80.68\% &  97.33\% &  92.05\% &  1.02\% \\
                                 & \citet{xiao2018training} L\tnote{*} &        166.39 &   37.45 &  203.84 &     59.6\% &  97.54\% &  93.25\% &  24.1\% \\
\cmidrule(lr){1-9}
\multirow{4}{4.5em}{CIFAR10 $\epsilon=\frac{2}{255}$} & EEV S &        0.0258 &  0.0013 &  0.0271 &    26.13\% &  46.58\% &  33.70\% &       0 \\
                                 & EEV L &        0.1653 &  0.0097 &  0.1750 &    30.49\% &  47.35\% &  38.22\% &       0 \\
                                 & \citet{xiao2018training} S &         52.58 &   13.50 &   66.08 &    45.93\% &  61.12\% &  49.92\% &  1.86\% \\
                                 & \citet{xiao2018training} L\tnote{*} &        335.97 &   29.88 &  365.85 &     41.4\% &  61.41\% &  50.61\% &   9.6\% \\
\cmidrule(lr){1-9}
\multirow{4}{4.5em}{CIFAR10 $\epsilon=\frac{8}{255}$} & EEV S &        0.0313 &  0.0014 &  0.0327 &    18.93\% &  37.75\% &  24.60\% &       0 \\
                                 & EEV L &        0.1691 &  0.0090 &  0.1781 &    22.55\% &  35.00\% &  26.41\% &       0 \\
                                 & \citet{xiao2018training} S &         38.34 &   22.33 &   60.67 &    20.27\% &  40.45\% &  26.78\% &  2.47\% \\
                                 & \citet{xiao2018training} L\tnote{*} &        401.72 &   20.14 &  421.86 &     19.8\% &  42.81\% &  28.69\% &   5.4\% \\
\bottomrule
\end{tabular}

        \begin{tablenotes}
        \item EEV is exact verification for BNNs with the proposed EEV system.
             \citet{xiao2018training} is exact verification for real-valued
            networks with data taken from~\cite{xiao2018training}. The
            \texttt{S} and \texttt{L} suffix indicates \texttt{conv-small} or
            \texttt{conv-large} architectures. The build time is the time
            required to generate the SAT or MILP formulation from the network
            weights and the input image. The solve time is the time that the
            solver spends solving the formulation. We limit the solving time for
            each case to $120$ seconds as in~\cite{xiao2018training}.
        \item[*] \citet{xiao2018training} test their large model only on the
            first $1000$ images due to slow verification.
        \end{tablenotes}
    \end{threeparttable}
\end{table}

\paragraph{Verifying a model ensemble with reject option:} We evaluate the
extensibility of our system by considering an ensemble of $M$ models that
rejects the input if they do not fully agree on the classification.  A
successful adversarial attack must present an input that causes all components
to produce the same wrong classification. The attack goal can be easily
formulated in CNF. Let $n$ be the number of classes, $c$ be the correct class,
and $r^{m}_{ij} = (y_i^m - y_j^m > 0)$ being a reified cardinality constraint
denote whether the score of class $i$ is higher than that of class $j$ in the
logits vector of model $m$. Let $f_i = \land_{1\le m \le M}\land_{1\le j\neq i
\le n}r^m_{ij}$ denote whether all models agree on class $i$. Then the attack
goal is simply $\lor_{1\le i \neq c \le n}f_i$. We present in
\cref{tab:ensemble} the results of verifying an ensemble of two models. The results show
that EEV can easily handle this complex query (i.e., $\argmax f_1(x) = \argmax
f_2(x) \neq c$) which is arguably nontrivial or even challenging to be
formulated as an efficiently solvable continuous optimization problem~\cite{
jonas2020certifying}, and that our solver is efficient in exploring the decision
space of a model ensemble.

\begin{table}[t]
    \scriptsize
    \centering
    \caption{Verifying a model ensemble with reject option on MNIST with a
        $\linf$ perturbation bound of $0.3$}
    \label{tab:ensemble}
    \vspace{1em}
    \begin{tabular}{lrrr}
\toprule
{} & Test Accuracy & Mean Solve Time (s) & Attack Success Rate \\
\midrule
\texttt{conv-small} &       94.31\% &               0.001 &             33.58\% \\
\texttt{conv-large} &       96.36\% &               0.004 &             22.41\% \\
ensemble            &       93.32\% &               0.003 &             17.21\% \\
\bottomrule
\end{tabular}

\end{table}


\section{Conclusion}

In this work we demonstrate that it is possible to significantly scale up the
exact verification of Binarized Neural Networks (BNNs) by equipping an
off-the-shelf SAT solver with domain-specific propagation rules and
simultaneously training solver-friendly robust BNNs. Although we focus on
verifying adversarial robustness, our method could be generalized to verify
other properties of BNNs. Our experimental results demonstrate the significant
performance gains that our techniques deliver.


\section*{Broader impact}

Binarized Neural Networks (BNNs) are attractive targets for deployment in a
variety of contexts including edge devices due to their efficiency advantages
over real-valued networks. The present research, by developing techniques for
training and verifying robust BNNs, may help enable the development of systems
that more reliably and predictably serve their goals --- systems that may be
less likely to exhibit unexpected behavior in response to new inputs; systems
that may be less vulnerable to attack. The research may therefore increase the
range and capabilities of systems that use BNNs. Because this technology is
general purpose and may be deployed in the service of prosocial, antisocial, or
mixed goals, the ultimate broader impacts may be shaped by the choices societies
make about how to use these capabilities. Example potential impacts include the
increased deployment of accurate surveillance systems, more accurate vision
systems for safer self-driving cars, more reliable autonomous control systems,
and less effort spent certifying systems that include BNNs.

\begin{ack}
    This work was supported by the Grass Instrument Company Fellowship in
    Electrical Engineering and the project ``Automatically Learning the Behavior
    of Computational Agents'' (MIT CO 6940111, sponsored by Boeing with sponsor
    ID \#Z0918-5060). We thank all the reviewers for providing the insightful
    comments that help further improve the quality of this paper, especially
    during this difficult year. Kai would like to thank Vijay Ganesh and Saeed
    Nejati for the helpful discussions on SAT solving, and Dimitris Tsipras for
    discussing training robust neural network. Kai also thanks Qi Song for her
    constant support during working on this project in the quarantine.
\end{ack}


\bibliographystyle{plainnat}
\bibliography{references.bib}

\appendix
\clearpage
\section{Details of BNN encoding}

We present details for encoding the inference computation of a single
linear-BatchNorm-binarize module in a BNN. Recall that such a module is defined
for an input $x\in\{0,1\}^n$, an output $y\in\{0,1\}^m$, and a weight $W \in
\real^{m\times n}$:
\begin{align}
    y = \bin_{act}(\bn(\bin_{w}(W) x)) \label{eqn:detail-bnn-layer-train}
\end{align}
where:
\begin{align*}
    \bin_{w}(W) &= \sgn(W) \in\{-1,1\}^{m\times n} \\
    \bn(x) &= \gamma \odot \frac{x-\expect[x]}{\sqrt{\var[x]+\epsilon}} + \beta \\
    & \text{\small \hspace{1em}
        is the Batch Normalization~\cite{ioffe2015batch} with
        parameters $\gamma\in\real^m$ and $\beta\in\real^m$} \\
    \bin_{act}(x) &= (x \ge 0) = (\sgn(x)+1)/2 \in\{0, 1\}^m
\end{align*}

Batch Normalization becomes a linear transformation in inference:
\begin{align}
    \xbn &=  \kbn \odot x + \bbn
\end{align}
where:
\begin{align*}
    \kbn &= \frac{\gamma}{\sqrt{\sigma^2+\epsilon}}  \\
    \bbn &= \beta - \kbn\mu \\
    \mu \hspace{1em} & \text{is the mean of $x$ on training set} \\
    \sigma^2 \hspace{1em} & \text{is the variance of $x$ on training set} \\
\end{align*}

With $\wbin = \bin_w(W)$ being a fixed parameter, we can rewrite the computation
of a single element of $y$ in \eqnref{detail-bnn-layer-train} as the following:

\begin{align}
    y_i = \left( \kbn_i \sum_{j=1}^n \wbin_{ij}x_j + \bbn_i \ge 0\right)
    \label{eqn:bnn-yi}
\end{align}

To encode \eqnref{bnn-yi} as a reified cardinality constraint, we consider $1$
and \texttt{TRUE} interchangeably and $0$ and \texttt{FALSE} interchangeably.
If $\wbin_{ij}=1$, we have $\wbin_{ij}x_j = x_j$, and if $\wbin_{ij}=-1$, we
rewrite $\wbin_{ij}x_j = -x_j = (1-x_j)-1 = \lnot x_j - 1$. With such
substitutions of $\wbin_{ij}x_j$ we obtain a reified cardinality constraint:
\begin{align}
    y_i &=  \left( \kbn_i \sum_{j=1}^n \wbin_{ij}x_j + \bbn_i \ge 0\right)
        \nonumber \\
        &=  \left( \kbn_i \left(\sum_{j=1}^n l_{ij}(x_j) + \bsat_i \right) +
            \bbn_i \ge 0\right)
        \nonumber \\
        &= \left( \sum_{j=1}^n l_{ij}(x_j) \gtreqless
            \left[b_i(\kbn,\wbin,\bbn)\right] \right)
\end{align}
where:
\begin{align*}
    l_{ij}(x_j) &= \left\{\begin{array}{ll}
            x_j & \text{ if } \wbin_{ij} = 1 \\
            0 & \text{ if } \wbin_{ij} = 0 \\
            \lnot x_j & \text{ if } \wbin_{ij} = -1 \\
    \end{array}\right. \\
    \bsat_i &= \sum_{j=1}^n \min\left(\wbin_{ij}, 0\right) \\
    b_i(\kbn,\wbin,\bbn) &= -\frac{\bbn_i}{\kbn_i} - \bsat_i \\
    \left(x \gtreqless y\right) &= \left\{\begin{array}{ll}
            x \ge y & \text{if $\kbn_i > 0$} \\
            x \le y & \text{if $\kbn_i < 0$}
    \end{array}\right. \\
    [x] &= \left\{\begin{array}{ll}
            \lceil x \rceil & \text{if $\kbn_i > 0$} \\
            \lfloor x \rfloor & \text{if $\kbn_i < 0$}
    \end{array}\right.
\end{align*}


\section{Experimental details}

\paragraph{Experimental environment} We conduct our experiments on a workstation
equipped with two GPUs (NVIDIA Titan RTX and NVIDIA GeForce RTX 2070 SUPER), 128
GiB of RAM and an AMD Ryzen Threadripper 2970WX 24-core processor. We use the
PyTorch~\cite{ NEURIPS2019_9015} framework to train all the networks.

\paragraph{Training method} We train the networks using the Adam
optimizer~\cite{kingma2014adam} for $200$ epochs with a minibatch size of $128$,
with the exception of the undefended \texttt{conv-small} networks on CIFAR10
which is trained for only $90$ epochs to avoid overfitting. Due to fluctuations
of test accuracy between epochs, we select from the last three epochs the model
with the highest natural test accuracy or PGD accuracy on the first $40$
training minibatches. The mean and variance statistics of batch normalization
layers are recomputed on the whole training set after training finishes.
Learning rate is initially $\scinum{1}{-4}$ and decayed to half on epoch $150$.
We use PGD with adaptive gradient cancelling to train robust networks, where the
perturbation bound $\epsilon$ is increased linearly from $0$ to the desired
value in the first $100$ epochs and the number of PGD iteration steps grows
linearly from $0$ to $10$ in the first $50$ epochs.

The parameter $\alpha$ in adaptive gradient cancelling is chosen to maximize the
PGD attack success rate evaluated on $40$ minibatches of training data sampled
at the first epoch. Candidate values of $\alpha$ are between $0.6$ to $3.0$ with
a step of $0.4$. Note that $\alpha$ is a global parameter shared by all neurons.

We do not use any data augmentation techniques for training. Due to limited
computing resource and significant differences between the settings we
considered, data in this paper are reported based on one evaluation run.

\paragraph{Weight initialization} All weights for the convolutional or fully
connected layers are initialized from a Gaussian distribution with standard
deviation $0.01$, and the mask weights $M_W$ in BinMask are enforced to be
positive by taking the absolute value during initialization.

\paragraph{Other hyperparameters} The input quantization step $s$ is set to be
$0.61$ for training on the MNIST dataset, and $0.064\approx16.3/255$ for
CIFAR10, which are chosen to be slightly greater than twice the largest
perturbation bound we consider for each dataset. The CBD loss is applied on
\texttt{conv-large} networks only and $\eta$ is set to be $\scinum{5}{-4}$
unless otherwise stated.  We apply a weight decay of $\scinum{1}{-7}$ on the
binarized mask weight $M_W$ of BinMask for \texttt{conv-small} and
\texttt{conv-large} networks, and the weight decay is $\scinum{1}{-5}$ for the
\texttt{MNIST-MLP} network. PGD accuracies reported for the test set are
evaluated with $100$ steps of PGD iterations.


\section{Adversarial robustness against varying perturbation bounds}

We run the verifier with varying perturbation bounds and present the time
in~\cref{tab:multi-eps-time} and the accuracy in~\cref{tab:multi-eps-acc}. The
bounds are set to be $\epsilon_0=0.1$, $\epsilon_1=0.2$, and $\epsilon_2=0.3$
for MNIST and $\epsilon_0=2/255$, $\epsilon_1=5/255$, and $\epsilon_2=8/255$
for CIFAR10. We note a few interesting discoveries:

\begin{enumerate}
    \item For the same network, the verification becomes slower for larger
        perturbation bounds. This behavior is expected because a larger bound
        corresponds to a larger input perturbation space, which makes the
        verification problem harder to solve.
    \item For the same bound used in verification, verification is faster for a
        more robust network that is trained with a larger perturbation bound.
        One plausible explanation is that robust networks are less sensitive to
        input changes, and they naturally allow the solver to learn simpler
        clauses that describe relationships between neurons. Another factor is
        that stronger adversarial training usually (but not always) induces more
        sparse weights as can be seen in \cref{tab:multi-eps-acc}.
    \item The gap between PGD accuracy and verifiable accuracy for a fixed test
        perturbation bound gets narrower as the training perturbation becomes
        stronger. More interestingly, when tested against perturbations of
        $\epsilon_0$, although the network has higher verifiable accuracy when
        it is trained with stronger perturbations, its PGD accuracy even gets
        lower (comparing PGD accuracy with $\epsilon=\epsilon_0$ in
        \cref{tab:multi-eps-acc}). Such phenomenon suggests that PGD accuracy is
        not always positively correlated with verfiable accuracy and the
        adversarial training algorithm could be further improved.
\end{enumerate}

\begin{table}[t]
    \caption{Verification time with varying perturbation bounds. Time limit is
        120 seconds.}
    \vspace{1em}
    \label{tab:multi-eps-time}
    \scriptsize
    \centering
    \begin{tabular}{llrrrrrrrrr}
\toprule
Dataset & Network & \multicolumn{3}{c}{Mean Solve Time} & \multicolumn{3}{c}{Solver Timeout} & \multicolumn{3}{c}{Mean Build+Solve Time} \\ \cmidrule(lr){3-5} \cmidrule(lr){6-8} \cmidrule(lr){9-11}
Training $\epsilon$&                     & $\epsilon=\epsilon_0$ & $\epsilon=\epsilon_1$ & $\epsilon=\epsilon_2$ & $\epsilon=\epsilon_0$ & $\epsilon=\epsilon_1$ & $\epsilon=\epsilon_2$ & $\epsilon=\epsilon_0$ & $\epsilon=\epsilon_1$ & $\epsilon=\epsilon_2$ \\
\midrule
\multirow{2}{5em}{MNIST $\epsilon=0.1$} & \texttt{conv-small} &                0.0004 &                0.0021 &                0.0713 &                     0 &                     0 &                0.01\% &                0.0162 &                0.0180 &                0.0876 \\
                         & \texttt{conv-large} &                0.0025 &                0.0129 &                0.1269 &                     0 &                     0 &                0.01\% &                0.1115 &                0.1197 &                0.2254 \\
\cmidrule(lr){1-11}
\multirow{2}{5em}{MNIST $\epsilon=0.3$} & \texttt{conv-small} &                0.0004 &                0.0004 &                0.0006 &                     0 &                     0 &                     0 &                0.0147 &                0.0155 &                0.0146 \\
                         & \texttt{conv-large} &                0.0010 &                0.0018 &                0.0039 &                     0 &                     0 &                     0 &                0.1162 &                0.1142 &                0.1179 \\
\cmidrule(lr){1-11}
\multirow{2}{5em}{CIFAR10 $\epsilon=2/255$} & \texttt{conv-small} &                0.0013 &                0.0017 &                0.0025 &                     0 &                     0 &                     0 &                0.0271 &                0.0298 &                0.0366 \\
                         & \texttt{conv-large} &                0.0097 &                0.0136 &                0.0141 &                     0 &                     0 &                     0 &                0.1750 &                0.1947 &                0.1918 \\
\cmidrule(lr){1-11}
\multirow{2}{5em}{CIFAR10 $\epsilon=8/255$} & \texttt{conv-small} &                0.0009 &                0.0011 &                0.0014 &                     0 &                     0 &                     0 &                0.0236 &                0.0284 &                0.0327 \\
                         & \texttt{conv-large} &                0.0087 &                0.0084 &                0.0090 &                     0 &                     0 &                     0 &                0.1704 &                0.1696 &                0.1781 \\
\bottomrule
\end{tabular}

\end{table}
\begin{table}[t]
    \caption{Verifiable accuracy with varying perturbation bounds.}
    \vspace{1em}
    \label{tab:multi-eps-acc}
    \scriptsize
    \centering
    \begin{tabular}{llrrrrrrrr}
\toprule
Dataset & Network & Test Accuracy & \multicolumn{3}{c}{PGD Adversarial Accuracy} & \multicolumn{3}{c}{Verifiable Accuracy} & Sparsity \\ \cmidrule(lr){4-6} \cmidrule(lr){7-9}
Training $\epsilon$&                     &             ~ &    $\epsilon=\epsilon_0$ & $\epsilon=\epsilon_1$ & $\epsilon=\epsilon_2$ & $\epsilon=\epsilon_0$ & $\epsilon=\epsilon_1$ & $\epsilon=\epsilon_2$ &        ~ \\
\midrule
\multirow{2}{5em}{MNIST $\epsilon=0.1$} & \texttt{conv-small} &       97.44\% &                  93.47\% &               86.22\% &               70.68\% &               89.29\% &               66.49\% &               25.45\% &     90\% \\
                         & \texttt{conv-large} &       97.46\% &                  95.47\% &               92.56\% &               86.98\% &               91.68\% &               75.35\% &               40.14\% &     91\% \\
\cmidrule(lr){1-10}
\multirow{2}{5em}{MNIST $\epsilon=0.3$} & \texttt{conv-small} &       94.31\% &                  91.74\% &               87.43\% &               80.70\% &               90.24\% &               82.14\% &               66.42\% &     94\% \\
                         & \texttt{conv-large} &       96.36\% &                  94.82\% &               92.19\% &               87.90\% &               93.71\% &               88.55\% &               77.59\% &     87\% \\
\cmidrule(lr){1-10}
\multirow{2}{5em}{CIFAR10 $\epsilon=2/255$} & \texttt{conv-small} &       46.58\% &                  33.70\% &               18.85\% &                9.32\% &               26.13\% &                8.26\% &                2.39\% &     94\% \\
                         & \texttt{conv-large} &       47.35\% &                  38.22\% &               28.20\% &               19.60\% &               30.49\% &               13.30\% &                4.98\% &     85\% \\
\cmidrule(lr){1-10}
\multirow{2}{5em}{CIFAR10 $\epsilon=8/255$} & \texttt{conv-small} &       37.75\% &                  33.88\% &               29.02\% &               24.60\% &               32.18\% &               24.82\% &               18.93\% &     96\% \\
                         & \texttt{conv-large} &       35.00\% &                  32.45\% &               29.17\% &               26.41\% &               31.20\% &               26.39\% &               22.55\% &     98\% \\
\bottomrule
\end{tabular}

\end{table}


\section{Ablation study}
\label{sec:ablation-study}

\begin{table*}[t]
    \scriptsize
    \center
    \caption{Comparison of methods on $40$ randomly chosen MNIST test images
        with solving time limit of $3600$ seconds.}
    \label{tab:cmp-method-mnist}
    \vspace{1em}
    \begin{adjustwidth}{-1in}{-1in}
        \begin{center}
        \bgroup
            \setlength{\tabcolsep}{1.3ex}
            \begin{tabular}{lL{14ex}L{17ex}R{11ex}R{9ex}lR{12ex}R{11ex}rR{12ex}}
\toprule
$\epsilon_{\text{train}}$ & Network Architecture & Training Method & Test Accuracy & Sparsity & Solver & Mean Solve Time & Median Solve Time & Timeout & Verifiable Accuracy \\
\midrule
\multirow{13}{*}{0} & \multirow{5}{*}{\texttt{conv-small}} & Ternary & 97.59\% & 81\% & MiniSatCS &         756.288 &             4.281 &    15\% &                 0\% \\
\cmidrule(lr){3-10}
\cmidrule(lr){4-10}
\cmidrule(lr){5-10}
    &                     & \multirow{4}{*}{BinMask} & \multirow{4}{*}{97.35\%} & \multirow{4}{*}{84\%} & MiniSatCS &           0.002 &             0.002 &       0 &                52\% \\
    &                     &             &         &      & MiniSat &           2.249 &             1.142 &       0 &                52\% \\
    &                     &             &         &      & Z3 &           0.089 &             0.089 &       0 &                52\% \\
    &                     &             &         &      & RoundingSat &           0.048 &             0.042 &       0 &                52\% \\
\cmidrule(lr){2-10}
\cmidrule(lr){3-10}
\cmidrule(lr){4-10}
\cmidrule(lr){5-10}
    & \multirow{8}{*}{\texttt{conv-large}} & Ternary & 99.07\% & 86\% & MiniSatCS &        2522.082 &          3600.002 &    68\% &                 0\% \\
\cmidrule(lr){3-10}
\cmidrule(lr){4-10}
\cmidrule(lr){5-10}
    &                     & Ternary+CBD & 95.58\% & 87\% & MiniSatCS &         886.007 &            21.711 &    20\% &                 0\% \\
\cmidrule(lr){3-10}
\cmidrule(lr){4-10}
\cmidrule(lr){5-10}
    &                     & Ternary+10xCBD & 92.91\% & 78\% & MiniSatCS &         342.097 &             4.742 &     5\% &                 2\% \\
\cmidrule(lr){3-10}
\cmidrule(lr){4-10}
\cmidrule(lr){5-10}
    &                     & BinMask & 98.94\% & 86\% & MiniSatCS &        2595.032 &          3600.001 &    70\% &                 2\% \\
\cmidrule(lr){3-10}
\cmidrule(lr){4-10}
\cmidrule(lr){5-10}
    &                     & \multirow{4}{*}{BinMask+CBD} & \multirow{4}{*}{96.88\%} & \multirow{4}{*}{89\%} & MiniSatCS &           0.664 &             0.028 &       0 &                70\% \\
    &                     &             &         &      & MiniSat &         225.861 &            18.761 &       0 &                70\% \\
    &                     &             &         &      & Z3 &         146.567 &             0.997 &       0 &                70\% \\
    &                     &             &         &      & RoundingSat &          33.922 &             0.702 &       0 &                70\% \\
\cmidrule(lr){1-10}
\cmidrule(lr){2-10}
\cmidrule(lr){3-10}
\cmidrule(lr){4-10}
\cmidrule(lr){5-10}
\multirow{18}{*}{0.3} & \multirow{7}{*}{\texttt{conv-small}} & Ternary (wd0) & 94.72\% & 80\% & MiniSatCS &         186.935 &             0.105 &     5\% &                30\% \\
\cmidrule(lr){3-10}
\cmidrule(lr){4-10}
\cmidrule(lr){5-10}
    &                     & Ternary (wd1) & 89.53\% & 93\% & MiniSatCS &           0.005 &             0.002 &       0 &                35\% \\
\cmidrule(lr){3-10}
\cmidrule(lr){4-10}
\cmidrule(lr){5-10}
    &                     & \multirow{5}{*}{BinMask} & \multirow{5}{*}{94.31\%} & \multirow{5}{*}{94\%} & MiniSatCS &           0.001 &             0.001 &       0 &                52\% \\
    &                     &             &         &      & MiniSat &           0.060 &             0.024 &       0 &                52\% \\
    &                     &             &         &      & Z3 &           0.040 &             0.040 &       0 &                52\% \\
    &                     &             &         &      & RoundingSat &           0.021 &             0.031 &       0 &                52\% \\
    &                     &             &         &      & MiniSat-CN &           0.034 &             0.008 &       0 &                52\% \\
\cmidrule(lr){2-10}
\cmidrule(lr){3-10}
\cmidrule(lr){4-10}
\cmidrule(lr){5-10}
    & \multirow{11}{*}{\texttt{conv-large}} & Ternary & 96.89\% & 91\% & MiniSatCS &        2828.479 &          3600.001 &    78\% &                 0\% \\
\cmidrule(lr){3-10}
\cmidrule(lr){4-10}
\cmidrule(lr){5-10}
    &                     & \multirow{4}{*}{Ternary+CBD} & \multirow{4}{*}{81.33\%} & \multirow{4}{*}{80\%} & MiniSatCS &           0.034 &             0.020 &       0 &                42\% \\
    &                     &             &         &      & MiniSat &         173.877 &            23.527 &       0 &                42\% \\
    &                     &             &         &      & Z3 &           2.093 &             1.840 &       0 &                42\% \\
    &                     &             &         &      & RoundingSat &           2.941 &             1.642 &       0 &                42\% \\
\cmidrule(lr){3-10}
\cmidrule(lr){4-10}
\cmidrule(lr){5-10}
    &                     & BinMask & 98.88\% & 82\% & MiniSatCS &        2442.698 &          3600.001 &    65\% &                 5\% \\
\cmidrule(lr){3-10}
\cmidrule(lr){4-10}
\cmidrule(lr){5-10}
    &                     & \multirow{5}{*}{BinMask+CBD} & \multirow{5}{*}{96.26\%} & \multirow{5}{*}{87\%} & MiniSatCS &           0.005 &             0.005 &       0 &                52\% \\
    &                     &             &         &      & MiniSat &           0.242 &             0.045 &       0 &                52\% \\
    &                     &             &         &      & Z3 &           0.530 &             0.540 &       0 &                52\% \\
    &                     &             &         &      & RoundingSat &           0.088 &             0.104 &       0 &                52\% \\
    &                     &             &         &      & MiniSat-CN &           0.388 &             0.076 &       0 &                52\% \\
\bottomrule
\end{tabular}

        \egroup
        \end{center}
    \end{adjustwidth}
\end{table*}
\begin{table*}[t]
    \scriptsize
    \center
    \caption{Comparison of methods on $40$ randomly chosen CIFAR10 test images
        with solving time limit of $3600$ seconds.}
    \label{tab:cmp-method-cifar10}
    \vspace{1em}
    \begin{adjustwidth}{-1in}{-1in}
        \begin{center}
        \bgroup
            \setlength{\tabcolsep}{1.3ex}
            \begin{tabular}{lL{14ex}L{17ex}R{11ex}R{9ex}lR{12ex}R{11ex}rR{12ex}}
\toprule
$\epsilon_{\text{train}}$ & Network Architecture & Training Method & Test Accuracy & Sparsity & Solver & Mean Solve Time & Median Solve Time & Timeout & Verifiable Accuracy \\
\midrule
\multirow{14}{*}{0} & \multirow{8}{*}{\texttt{conv-small}} & \multirow{4}{*}{Ternary} & \multirow{4}{*}{54.78\%} & \multirow{4}{*}{82\%} & MiniSatCS &           0.267 &             0.006 &       0 &                 0\% \\
        &                     &             &         &      & MiniSat &         327.303 &            50.036 &     7\% &                 0\% \\
        &                     &             &         &      & Z3 &         411.638 &           117.604 &     5\% &                 0\% \\
        &                     &             &         &      & RoundingSat &           0.361 &             0.098 &       0 &                 0\% \\
\cmidrule(lr){3-10}
\cmidrule(lr){4-10}
\cmidrule(lr){5-10}
        &                     & \multirow{4}{*}{BinMask} & \multirow{4}{*}{55.22\%} & \multirow{4}{*}{79\%} & MiniSatCS &           0.003 &             0.003 &       0 &                 0\% \\
        &                     &             &         &      & MiniSat &           3.981 &             3.577 &       0 &                 0\% \\
        &                     &             &         &      & Z3 &           0.590 &             0.376 &       0 &                 0\% \\
        &                     &             &         &      & RoundingSat &           0.081 &             0.077 &       0 &                 0\% \\
\cmidrule(lr){2-10}
\cmidrule(lr){3-10}
\cmidrule(lr){4-10}
\cmidrule(lr){5-10}
        & \multirow{6}{*}{\texttt{conv-large}} & Ternary & 69.25\% & 89\% & MiniSatCS &         823.370 &             1.860 &    20\% &                 0\% \\
\cmidrule(lr){3-10}
\cmidrule(lr){4-10}
\cmidrule(lr){5-10}
        &                     & BinMask & 67.46\% & 94\% & MiniSatCS &         300.404 &             3.201 &     5\% &                 0\% \\
\cmidrule(lr){3-10}
\cmidrule(lr){4-10}
\cmidrule(lr){5-10}
        &                     & \multirow{4}{*}{BinMask+CBD} & \multirow{4}{*}{63.18\%} & \multirow{4}{*}{88\%} & MiniSatCS &           1.415 &             0.048 &       0 &                 0\% \\
        &                     &             &         &      & MiniSat &         168.079 &            69.471 &       0 &                 0\% \\
        &                     &             &         &      & Z3 &        3515.386 &          3600.121 &    92\% &                 0\% \\
        &                     &             &         &      & RoundingSat &          19.162 &             0.858 &       0 &                 0\% \\
\cmidrule(lr){1-10}
\cmidrule(lr){2-10}
\cmidrule(lr){3-10}
\cmidrule(lr){4-10}
\cmidrule(lr){5-10}
\multirow{13}{*}{$8/255$} & \multirow{5}{*}{\texttt{conv-small}} & Ternary & 32.59\% & 95\% & MiniSatCS &           0.002 &             0.002 &       0 &                15\% \\
\cmidrule(lr){3-10}
\cmidrule(lr){4-10}
\cmidrule(lr){5-10}
        &                     & \multirow{4}{*}{BinMask} & \multirow{4}{*}{37.75\%} & \multirow{4}{*}{96\%} & MiniSatCS &           0.001 &             0.002 &       0 &                18\% \\
        &                     &             &         &      & MiniSat &           0.070 &             0.082 &       0 &                18\% \\
        &                     &             &         &      & Z3 &           0.050 &             0.052 &       0 &                18\% \\
        &                     &             &         &      & RoundingSat &           0.033 &             0.043 &       0 &                18\% \\
\cmidrule(lr){2-10}
\cmidrule(lr){3-10}
\cmidrule(lr){4-10}
\cmidrule(lr){5-10}
        & \multirow{8}{*}{\texttt{conv-large}} & \multirow{2}{*}{Ternary} & \multirow{2}{*}{34.60\%} & \multirow{2}{*}{94\%} & MiniSatCS &         241.572 &             0.047 &     5\% &                10\% \\
        &                     &             &         &      & RoundingSat &         516.767 &             1.154 &    12\% &                10\% \\
\cmidrule(lr){3-10}
\cmidrule(lr){4-10}
\cmidrule(lr){5-10}
        &                     & \multirow{2}{*}{BinMask} & \multirow{2}{*}{53.91\%} & \multirow{2}{*}{88\%} & MiniSatCS &         206.037 &             0.052 &     5\% &                 0\% \\
        &                     &             &         &      & RoundingSat &         301.062 &             0.546 &     7\% &                 0\% \\
\cmidrule(lr){3-10}
\cmidrule(lr){4-10}
\cmidrule(lr){5-10}
        &                     & \multirow{4}{*}{BinMask+CBD} & \multirow{4}{*}{38.75\%} & \multirow{4}{*}{87\%} & MiniSatCS &           0.009 &             0.011 &       0 &                20\% \\
        &                     &             &         &      & MiniSat &           0.224 &             0.267 &       0 &                20\% \\
        &                     &             &         &      & Z3 &           0.768 &             0.795 &       0 &                20\% \\
        &                     &             &         &      & RoundingSat &           0.218 &             0.126 &       0 &                20\% \\
\bottomrule
\end{tabular}

        \egroup
        \end{center}
    \end{adjustwidth}
\end{table*}

We conduct comprehensive experiments to study the effectiveness of our proposed
methods --- namely BinMask, CBD loss, and native handling of the reified
cardinality constraints in the \texttt{MiniSatCS} verifier --- under different
settings. We apply various combinations of ternary quantization, BinMask, and
CBD loss during training, and verify the networks using multiple solvers. The
experimental results are presented in \cref{tab:cmp-method-mnist} and
\cref{tab:cmp-method-cifar10}.

For each dataset, we train the \texttt{conv-small} and \texttt{conv-large}
networks under two training settings: undefended (i.e., $\epsilon=0$) and
adversarial training with a large perturbation bound ($\epsilon = 0.3$ for MNIST
and $\epsilon = 8/255$ for CFAR10). We tune the weight decay coefficient of
ternary quantization so that the total sparsity is close to that of BinMask or
BinMask+CBD. All BinMask networks have the same weight decay of
$\scinum{1}{-7}$, except for the undefended \texttt{conv-large} networks on
CIFAR10 which have a larger weight decay of $\scinum{2.5}{-6}$ due to the low
sparsity under the default setting.

\newcommand{\satz}{\texttt{Z3}}
\newcommand{\satm}{\texttt{MiniSat}}
\newcommand{\satmcn}{\texttt{MiniSat-CN}}
\newcommand{\satmcs}{\texttt{MiniSatCS}}
\newcommand{\satpb}{\texttt{RoundingSat}}

We consider the following questions for ablation study:

\begin{qanda}
    \Q Does native handling of reified cardinality constraints always facilitate
        the SAT solving?
    \A Yes. We compare the solving time of \satmcs, \satm, \satz, and \satpb~on
        both network architectures trained on both datasets. The sequential
        counters encoding is used for \satm, and we also evaluate \satmcn~that
        uses the cardinality networks encoding on a few cases, but it is not
        consistently better than \satm. Our solver \satmcs, extended from
        \satm~with native handling of reified cardinality constraints, delivers
        a speedup of mean solving time by a factor of between \mcsSpeedupMin~to
        \mcsSpeedupMax~times compared all other solvers in all cases, and the
        average speedup is \mcsSpeedupAvg. The speedup of median solving time of
        \satmcs~compared to others is \mcsSpeedupMidMin~ to \mcsSpeedupMidMax.
        Note that although the encoding complexity of \satmcn~is $O(n\log^2b)$
        which is asymptotically better than $O(nb)$ of \satm, the low
        cardinality bounds in our networks make such asymptotic comparison
        inaccurate. Also note that \satmcs~is constantly faster than all other
        solvers, and no solver is constantly the second fastest (\satpb~is
        usually faster than \satz~and \satm, but it is slower than \satz~in the
        \texttt{Ternary+CBD} setting with the adversarially trained
        \texttt{conv-large} network on MNIST).

    \Q How fast do BinMask networks verify compared to ternary quantization
        networks?
    \A For the \texttt{conv-small} networks, BinMask networks verify
        significantly faster than ternary networks with similar
        sparsity, especially in the undefended training setting. For the
        \texttt{conv-large} networks without the CBD loss, neither of them
        constantly verifies faster than the other. We note that for the
        \texttt{conv-large} networks, BinMask still produces more balanced
        layer-wise sparsities, which are, for example, $[\layerSparsityTernary]$
        and $[\layerSparsityBinMask]$ for the undefended ternary and BinMask
        networks on MNIST respectively. Their verification speeds are both slow
        because the high cardinality bounds dominate verification complexity,
        which are \layerSparsityTernaryCardBound~and
        \layerSparsityBinMaskCardBound~on average for each neuron in the two
        networks respectively.

    \Q How accurate are BinMask networks compared to ternary quantization
        networks?
    \A Interestingly, ternary quantization networks have slightly higher test
        accuracy in most of the undefended training cases, but BinMask networks
        have both higher test accuracy and verifiable accuracy when trained
        adversarially.  We highlight the comparison of \texttt{Ternary (wd0)},
        \texttt{Ternary (wd1)}, and \texttt{BinMask} on the adversarially
        trained \texttt{conv-small} network in \cref{tab:cmp-method-mnist}. The
        \texttt{wd0} ternary network has a weaker weight decay to match the
        accuracy with the BinMask network, but it verifies much slower and has
        lowered verifiable accuracy. The \texttt{wd1} ternary network is trained
        with a stronger weight decay to match the sparsity of the BinMask
        network, but it has much lower test accuracy and also verifies slower.
        Our results suggest that BinMask not only reduces verification
        complexity, but also regularizes model capacity to make it more robust.

    \Q Does the CBD loss reduce cardinality bound and speed up verification for
        ternary quantization networks?
    \A Yes, but less effectively. We train \texttt{conv-large} networks on
        MNIST with ternary quantization and CBD loss as the \texttt{Ternary+CBD}
        networks shown in \cref{tab:cmp-method-mnist}. The CBD loss induces
        denser networks with ternary quantization, and the weight decay of
        \texttt{Ternary+CBD} networks is increased by five times compared to
        \texttt{Ternary} networks to maintain comparable sparsity. The average
        cardinality bound of the undefended \texttt{Ternary+CBD} network is
        \ternWeightCbdCardBound, compared to \ternWeightCardBound~of the
        \texttt{Ternary} network. Although the \texttt{Ternary+CBD} and
        \texttt{BinMask+CBD} networks achieve similar average cardinality bounds
        (\ternWeightCbdCardBound~vs \ternWeightBinMaskCbdCardBound), the ternary
        network has much higher maximal cardinality bound (%
        \ternWeightCbdCardBoundMax~vs \ternWeightBinMaskCbdCardBoundMax).
        Therefore, the verification time of \texttt{Ternary+CBD} is
        significantly improved over \texttt{Ternary} but is still longer than
        that of the \texttt{BinMask+CBD} network. \texttt{Ternary+10xCBD} is
        obtained by increasing the CBD loss coefficient $\eta$ by ten times on
        \texttt{Ternary+CBD}, which has a lower average cardinality bound of
        \ternWeightStrongCbdCardBound, but the maximal cardinality bound is not
        decreased (\ternWeightStrongCbdCardBoundMax) and its test accuracy is
        much worse. The adversarially trained \texttt{Ternary+CBD} network has a
        lower maximal cardinality bound of \ternWeightAdvCbdCardBoundMax, which
        also verifies faster.  Ternary quantization networks with the CBD loss
        suffer from a larger decline of test accuracy compared to BinMask
        networks. Our results suggest that lower cardinality bound reduces
        verification complexity, and the BinMask formulation makes it easier to
        optimize for lower cardinality bounds.

    \Q Do other solvers benefit from more balanced layer-wise sparsities and/or
        lower cardinality bounds?
    \A Yes. We evaluate \satm, \satz, and \satpb~on a relatively easy to verify
        ternary network (undefended \texttt{conv-small} on CIFAR10 in
        \cref{tab:cmp-method-cifar10}).  The results show that all the solvers
        achieve significant speedup on the corresponding BinMask network,
        although \satz~benefits more from BinMask than \satm~and \satpb. We also
        try \satm~and \satz~on the easiest-to-verify \texttt{conv-large}
        network trained with only BinMask (i.e., the one adversarially trained
        on CIFAR10), but \satm~fails due to out of memory error, and
        \satz~always exceeds the one hour time limit (data not shown in the
        table).  With \texttt{BinMask+CBD} they both verify much faster.
        \satpb~also benefits from \texttt{BinMask} and \texttt{BinMask+CBD} in
        this setting as shown in the table.  Note that the ternary network has
        lower test accuracy and higher overall sparsity, but still verifies
        slower for all the solvers. Our results suggest that our two strategies,
        which are inducing more balanced layer-wise sparsity and lower
        cardinality bounds, both reduce the complexity of the verification
        problem and facilitate all the solvers that we have considered.
\end{qanda}


\end{document}